%% file: main.tex
\documentclass[9pt,twocolumn,twoside]{pnas-new}

\usepackage{lipsum}

\templatetype{pnasresearcharticle} 

\title{Self-supervised Pretraining for Partial Differential Equations}

\author[a, 1]{Varun Madhavan}
\author[b, 1]{Amal S Sebastian} 
\author[c]{Bharath Ramsundar}
\author[b,d, 2]{Venkatasubramanian Viswanathan}

\affil[a]{Indian Institute of Technology Kharagpur, Kharagpur, West Bengal 721302, India}
\affil[b]{Department of Aerospace Engineering, University of Michigan, Ann Arbor, Michigan 48109, USA}
\affil[c]{Deep Forest Sciences, Palo Alto, California 94306, USA}
\affil[d]{Department of Mechanical Engineering, Carnegie Mellon University, Pittsburgh,
Pennsylvania 15213, USA}

\leadauthor{Madhavan} 

\equalauthors{\textsuperscript{1}equal contribution}
\correspondingauthor{\textsuperscript{2}E-mail: venkvis@umich.edu}

\begin{abstract}
\input{src/abstract}
\end{abstract}

\doi{\url{www.pnas.org/cgi/doi/10.1073/pnas.XXXXXXXXXX}}

\begin{document}

\maketitle
\ifthenelse{\boolean{shortarticle}}{\ifthenelse{\boolean{singlecolumn}}{\abscontentformatted}{\abscontent}}{}



\input{src/introduction}


\input{src/methodology/content}

\input{src/experiments/content}

\input{src/conclusion}

\showmatmethods{} 


\showacknow{} 

\clearpage \bibliography{refs}

\appendix

\input{src/supplementary}

\end{document}

%% file: src/abstract.tex
In this work, we describe a novel approach to building a neural PDE solver leveraging recent advances in transformer based neural network architectures. Our model can provide solutions for different values of PDE parameters without any need for retraining the network. The training is carried out in a self-supervised manner, similar to pre-training approaches applied in language  and vision tasks. We hypothesize that the model is in effect learning a family of operators (for multiple parameters) mapping the initial condition to the solution of the PDE at any future time step $t$. We compare this approach with the Fourier Neural Operator (FNO), and demonstrate that it can generalize over the space of PDE parameters, despite having a higher prediction error for individual parameter values compared to the FNO. We show that performance on a specific parameter can be improved by finetuning the model with very small amounts of data. We also demonstrate that the model scales with data as well as model size.

%% file: src/introduction.tex
\section*{Introduction}
    PDEs are ubiquitous when modelling physical phenomena. Over the past few decades, significant progress has been made towards being able to numerically solve PDEs for various scientific and engineering use cases like climate modelling \cite{LYNCH20083431}, aircraft design \cite{Keane2005}, and geophysical electromagnetics \cite{Haber2014}.
    
    Traditional numerical methods can be very computationally expensive for real time and large scale applications due to a myriad of reasons such as the size of the discretized system, the complexity of the physics, to the choice of solver. Carrying out design based tasks that require multiple simulations of the model by systematically perturbing the system parameters becomes unfeasible as a change of parameters necessitates restarting the simulation from scratch.

    Recently machine learning based PDE solvers have been gaining a lot of interest \cite{Sirignano2018, Zang2020, Raissi2019}. These solvers can overcome the bottlenecks of traditional numerical methods, as they are able to produce solutions of similar accuracy but in a fraction of the time.
    
    The idea of learning the solution operator using a machine learning based model \cite{NO2023} has been receiving a lot of attention. The learned solution operator can be a finite-dimensional operator \cite{FDO-1, FDO-2, FDO-3} that maps between finite dimensional Euclidean spaces or it can be classified as a neural operator \cite{NeuralOperator, DeepONet, NO-2} that learns mappings between function spaces. Finite dimensional operators are by definition mesh dependent whereas neural operators are grid invariant \cite{FNO}. Out of the many architectures used to build neural solvers, the Fourier Neural Operator (FNO) \cite{FNO} has shown to outperform most other deep learning methods over a variety of different systems \cite{PDEBench}. 

    While significant progress has been made towards building fast and efficient neural solvers, the majority of works in the field are still limited by the fact that the models are trained to solve for a fixed value of PDE parameters. These models are not amenable to produce accurate solutions to parameters outside the training distribution without retraining the network from scratch. While some attempts have been made to solve this problem \cite{brandstetter2022message, cape, FMfluid}, none of them use a transformer based architecture. While \cite{ViTOperator} has used a Vision transformer to create a neural solver it suffers from the problem of only being applicable to a fixed set of PDE parameters.
    
    Transformers are known to demonstrate strong scaling properties with data and model size \cite{gpt1, gpt2, gpt3}, making them the ideal candidate to build large foundation models. The success of large language models (LLM's) can be attributed to the use of large pre-trained models or foundation models \cite{HAN2021225}. These models trained on large amounts of unlabelled data and are able to learn and store very rich information in their parameters. This allows for very quick retraining of the model with significantly lower amounts of data when dealing with specific downstream tasks. This paradigm of pretraining and finetuning has been adopted for various use cases like molecular property prediction \cite{chemberta, molformer}, predicting evolutionary dynamics of viruses \cite{genslm}, weather and climate forecasting \cite{climax} with great success. 
    
    The ability to generalize across PDE parameters is required for a scientific foundation model \cite{polymathic}, and thus motivates the choice of architecture. In particular, we choose to make use of a vision transformer \cite{vit}. We make use of the notion of pre-training and fine-tuning to build a neural PDE solver that can be quickly retrained to produce accurate solutions for new parameters with very low amounts of data. This is achieved by building a base neural solver pretrained on a large amounts of data from multiple PDE parameter values and then finetuned with a small amounts of data from the specific PDE parameter of interest.
    
     We use an approach similar to \cite{vit}, where we treat the PDE solution at different timesteps as an image where the channels of the image now contains information relevant to the system such as the state, the PDE parameter, and time. We train the network by providing the model with a context window of $n$ timesteps and make it predict the next timestep. However, during validation and testing we carry out an auto regressive roll-out of the solution.

    We show results from experiments on multiple systems to compare performance against an FNO, and show performance scaling with more data and varying model sizes. Experiments are carried out on the one dimensional advection equation, one dimensional viscous Burgers equation and the two dimensional compressible Navier Stokes equation.


%% file: src/methodology/content.tex
\section*{Methodology}

\input{src/methodology/background}


\input{src/methodology/problem-formulation}

\subsection*{Approach}
    \label{subsec:approach}

    We use a modified Vision Transformer to learn $\mathbb{G}^{*}_{\Omega, \delta}$. A PDE solution $f(\bar{x}, t)$ is realised as a spatio-temporal grid  where
    \begin{equation*}
     \bar{x} = \{\bar{x}_1, \bar{x}_2, ... \bar{x}_N \} \in X \subset \mathbb{R}^k
    \end{equation*}
    where $k = 1, 2, 3$ for 1D, 2D, and 3D problems respectively, and
    \begin{equation*}
        t = \{t_1, t_2, ... t_T \} \in T
    \end{equation*} is the temporal discretization.
    
    The network takes as input the context window $\{ f(\bar{x}, t_{i-k}), f(\bar{x}, t_{i-(k-1)}), ..., f(\bar{x}, t_{i-1}) \}$ to predict $\hat{f}(x, t_i)$, as shown in Figure \ref{figure:vit}. PDE parameter information for a given context window is added in as an additional tensor dimension with the context window.

    
    This input is converted into patches via a convolution operation. These patches are then projected, along with position embeddings, into patch embeddings of dimension [$h$, 1], where $h$ is the hidden size of the transformer. The output of the transformer is then converted back into patches via a linear projection. We recover a mesh from the output patches, via a deconvolution operation, which after a linear projection gives us our final prediction $\hat{f}(x, t_i)$.

    
    The weights of the network are learnt using the Adam optimizer, minimizing the objective described in Equation \ref{eq:training-loss}.

    For a given PDE, we train the network on a subset of the parameters $\omega_{\text{pt}} \in \{\omega_1, \omega_2, \dots, \omega_n\} \subset \Omega$ from the dataset. The testing dataset is divided into two parts - 1) the "in-domain" (id) set containing unseen trajectories from parameter values present in the training set, and 2) the "out-of-domain" (ood) set containing trajectories belonging to parameter values not present in the training set. This split is done to measure the performance of the learned networks on new parameter values.  
    \begin{figure}[h]
        \centering
        \includegraphics[width=\linewidth]{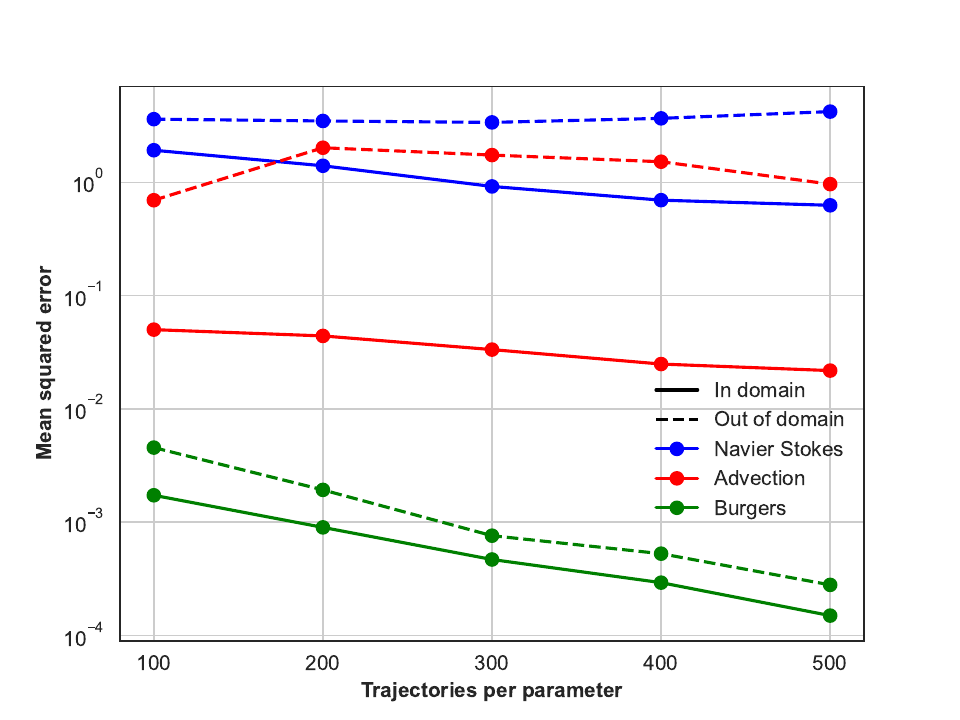}
        \caption{\textbf{In domain and out of domain scaling with more data}: The above plot shows how the prediction MSE varies with an increase in the training dataset size (i.e. the number of trajectories per parameter value $\omega$). In the in domain setting we observe that the prediction error decreases steadily with an increase in the dataset size for all systems. In the out of domain setting for Burgers, we observe that the prediction error decreases steadily with an increase in the dataset size. For advection, the prediction error decreases slowly from 200 trajectories/$\omega$ onwards.  The models could be under-fitting  leading do this trend. For the 2D-NSE, however, no discernible trend is observed. This again indicates that the models are possibly under-fitting the training sets, thus adding more data does not improve performance.}
        \label{fig:scaling-data-id}
    \end{figure}

\input{src/methodology/data}

%% file: src/methodology/background.tex
\subsection*{Background}        


    \subsubsection*{Transformer}
        \label{section:Transformers}


        The Transformer architecture \cite{attention} is made up of a stack of $L$ Transformer encoder layers. As shown in Figure \ref{figure:vit}, each Transformer encoder consists of multi-headed self-attention and MLP (i.e. feed-forward) blocks, making use of layer normalization \cite{layernorm} and residual connections \cite{residual_connections}. The self attention layer enables the model to weigh the importance of different tokens against each other and pick up on both global and local context, thus allowing the model to determine how much focus it needs to put on different parts of the input.


    \subsubsection*{Vision Transformer (ViT)}
        \label{section:ViT}
    
\begin{figure*}[t] 
            \centering
            \begin{minipage}{0.95\textwidth}
                \includegraphics[width=\linewidth]{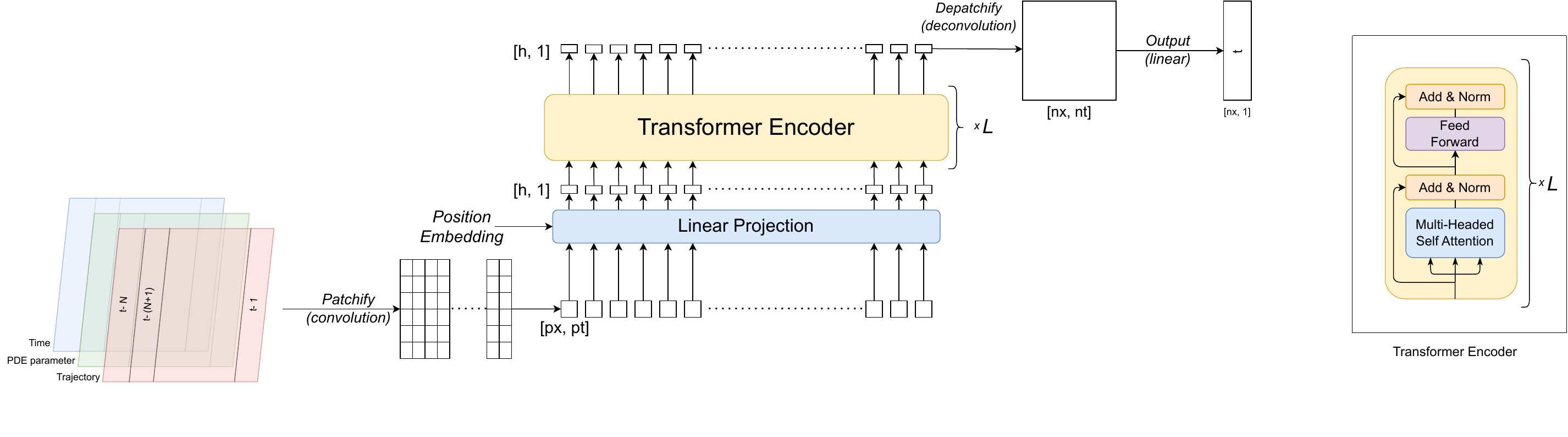}
                \caption{\textbf{PDE Transformer}: The fluid transformer works by treating the context window of $N$ time steps as an image with channels containing information about the PDE parameters and time. This image is passed through a convolution layer to break the image into patches of dimension $[\text{px, pt}]$ and the projected into a tokens of size $h$. After passing through L transformer encoder layers (shown on the right) we reconstruct the image through a transposed convolution, from which we extract the state at time $N+1$.}
                \label{figure:vit}
            \end{minipage}\hfill 
        \end{figure*}

        The Vision Transformer \cite{vit} (Figure \ref{figure:vit}) modifies the transformer \cite{attention} architecture to carry out image related tasks. This is achieved by breaking the input image into \textit{patches}. A 2 dimensional image of size $l_i \times b_i$ is converted to a sequence of $ \mathrm(l_i/l_p) \times \mathrm(b_i/b_p) $ patches of size $l_p \times b_p$ using a convolution layer with a kernel of size $l_p \times b_p$ and a stride length equal to $l_p \times b_p$.

        Each of these patches is then projected via a feed-forward layer to embedding vectors of a fixed shape. This sequence is then passed to the Transformer, which applies multi-headed self-attention to the sequence to learn context information to each embedding, i.e. information about the other relevant patches that it must "attend to". We extend this idea of patching the input "image" to PDE solutions in the approach subsection.  
        
        

%% file: src/methodology/problem-formulation.tex
\subsection*{Problem Formulation} 

    Let $f: \mathbb{R}^{n} \times \mathbb{R} \mapsto \mathbb{R}^{m}$, be the solution of the PDE we wish to solve. We define a mapping $\mathrm{G}_{\omega}$:
    \begin{equation}
        \mathrm{G}_{\omega} : \mathrm{A} \mapsto \mathrm{U}
    \end{equation}
    where $\mathrm{A}(\mathbb{R}^{d_a})$ is a Banach space that contains the initial condition, and $\mathrm{U}(\mathbb{R}^{d_u}$) is a Banach space containing the solution of the PDE for a specified set of PDE parameter values denoted as $\omega \in \Omega$.

    We solve for the family of operators $\mathbb{G}_{\Omega}$, that maps the initial condition to the solution of the PDE at time t, for any value of the PDE parameters:
    
    \begin{equation}
        \mathbb{G}_{\Omega}: f(., 0) \mapsto f(., t)
    \end{equation}

    This is realized by treating the family of operators as a time stepping operation of a function $f$, i.e:
    
    \begin{equation}
        \mathbb{G}_{\Omega, \delta}: f(\cdot, t) \mapsto f(\cdot, t + \delta t)
    \end{equation}

    We can recover $\mathbb{G}_{\Omega}$ as $\mathbb{G}_{\Omega} = \mathbb{G}_{\Omega, \delta}(\mathbb{G}_{\Omega, \delta}(\mathbb{G}_{\Omega, \delta}(\dots)))$. We attempt to learn an approximation of the operator $\mathbb{G}_{\Omega, \delta}$, by creating a parametric map:
    
    \begin{equation}
        \mathbb{G}^{*}_{\Omega, \delta}: f(\cdot, t) \times \Theta \mapsto f(\cdot, t + \delta t)
    \end{equation}
    
    where $\Theta$, is some finite dimensional parameter space.

    We do so by minimizing the following objective:
    
    \begin{equation}
        \sum_{\Omega}\sum_{t}\sum_{n}\lVert \mathbb{G}_{\Omega, \delta}^{*}(u_n) - \mathbb{G}_{\Omega, \delta}(u_n) \rVert_2 ^ 2
        \label{eq:training-loss}
    \end{equation}

    where $u_n$ is the solution of the nth initial condition at time t.

%% file: src/methodology/data.tex
\subsection*{Data}

\input{src/experiments/tables/baselines}

We use trajectory data from PDEBench \cite{PDEBench} to train our network. We refer to trajectory as the time evolution of a system from the initial condition to the solution at a given time $t$. The PDEBench dataset includes a large number of trajectories for multiple systems, with each trajectory starting from a unique initial condition. Further, data is available over a large parameter space for each system, which makes it ideal for the problem at hand.

For any given value in the parameter space, we have trajectories starting from the same set of initial conditions. This allows the model to learn the evolution of an initial condition under various parameter values. While PDEBench offers data for a large number of systems, we limit our analysis to the following PDEs:

\begin{itemize}
    \item 1D linear advection, where the wave speed $a$, is the parameter of interest.
    \item 1D Burgers equation, where the diffusion coefficient $\nu$, is the parameter of interest.
    \item 2D Compressible Navier Stokes, where the Mach number ($M$), shear ($\eta$) and bulk($\zeta$) viscosities are the parameters of interest. 
\end{itemize}

More details about the equations can be found in the Supplementary information at Section \ref{subsec:data_appendix}.

Henceforth we shall refer to our model as the PDE Transformer, which we abbreviate to PDE\kern+.15em-T.

%% file: src/experiments/tables/baselines.tex
\newcounter{one}
\setcounter{one}{1}

\newcounter{two}
\setcounter{two}{2}

\begin{table*}[ht!]
    \centering
    \begin{tabular}{@{}lcc|cc|cc@{}}
    \toprule
    & \multicolumn{2}{c}{\textbf{Advection}} & \multicolumn{2}{|c|}{\textbf{Burgers}} & \multicolumn{2}{c}{\textbf{Navier Stokes}} \\ \midrule
    & id-mean & ood-mean & id-mean & ood-mean & id-mean & ood-mean \\
    FNO [\Roman{one}] & 1.08E-04 & - & 7.72E-02 & - & 2.96E-04 & - \\ 
    FNO [\Roman{two}] & 1.02E-04 & - & 8.40E-06 & - &  3.26E-04& - \\
    ViT & 3.73E-03 & 3.61E-01 & 6.16E-04 & 1.82E-03 & 2.81E-01 & 8.90E-00 \\ \bottomrule
    \end{tabular}%
    \caption{Comparisons: The FT models are trained on 200 trajectories for each parameter value. FNO [\Roman{one}] refers to FNO models trained on 200 trajectories averaged over each parameter value. FNO [\Roman{two}] refers to FNO models trained on an equal number of trajectories as the corresponding ViT model (i.e. 200 * number of unique parameter values for each system).}
    \label{table:comparisons}
\end{table*}

%% file: src/experiments/content.tex
\section*{Experiments}
    \begin{figure}[]
        \centering
        \includegraphics[width=\linewidth]{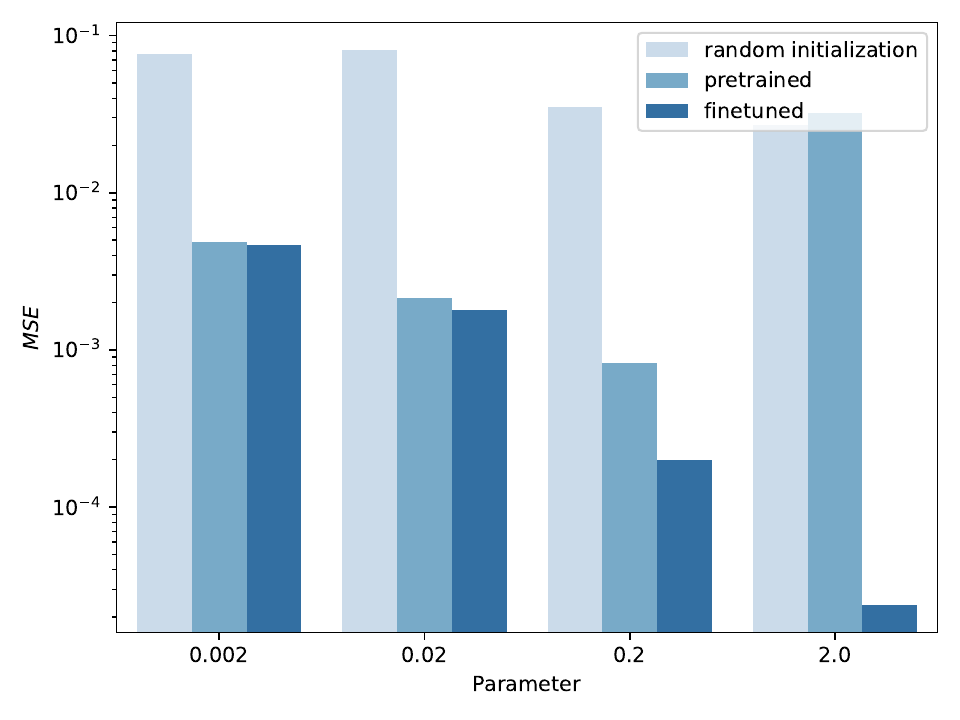}
        \caption{\textbf{Finetuning a pretrained model}: Finetuning a pretrained model for the Burgers equation on the out of domain parameters helps to improve performance for most parameter values. Not that for most parameter values, the pretrained model substantially outperforms random initialization.}
        \label{fig:finetuning}
    \end{figure}

\subsection*{Comparison with the FNO}
    
    We evaluate the performance of PDE\kern+.15em-T across multiple parameters, along with the FNO trained on a single parameter in table \ref{table:comparisons}. Performance is measured in terms of the mean squared error (MSE) between the predicted trajectory and the actual trajectory (averaged across all dependent variables). "id-mean" refers to the mean MSE across all trajectories in the in-domain test set, while "ood-mean" refers to the same measure for the out-of-domain test set. 

    The values of the parameters used in the in domain and out of domain sets are given is Table \ref{table:parameters}.
    
    
    

    A direct one-to-one comparison between the two models is not possible as the PDE\kern+.15em-T sees trajectories from various parameter values whereas the FNO is trained on a single parameter value. Thus to create a fair comparison we train two variants of the FNO. In this experiment we train the PDE\kern+.15em-T using 200 trajectories from each parameter listed in Table \ref{table:parameters}. FNO[I] is only trained using the 200 trajectories from one parameter, i.e FNO[I] is trained on subset of the training data used for the PDE\kern+.15em-T. Thus we also train another variant, FNO[II], which is trained on the same number of trajectories (but only from one parameter) as the PDE\kern+.15em-T i.e the size of the training data is same for both PDE\kern+.15em-T and FNO[II].

    While our approach has higher errors on average we note that, unlike the FNO, the PDE\kern+.15em-T can generalize to a wide range of parameter values. Performance gains can be made for a specific set of parameter values via fine-tuning which we shall demonstrate in the next section.
    
    

\subsection*{Does training with a few trajectories for the unseen parameters help?} 
    \label{section:finetuning}
    
    \input{src/experiments/tables/parameters}
    As we see in Table \ref{table:comparisons}, the PDE\kern+.15em-T is able to generalize to trajectories with out-of-domain parameter values. However, as one would expect, the errors are higher for the OOD parameters. Previous work applying transformers to the language and vision domains has demonstrated that performance on "new" tasks, analogous to new parameter values in our case, can be increased by further training ("fine-tuning") the model on a relatively small number of examples (as compared to the number of training examples; in our context, the number of trajectories per parameter during training) from that task.

        \begin{figure}
        \centering
        \includegraphics[width=\linewidth]{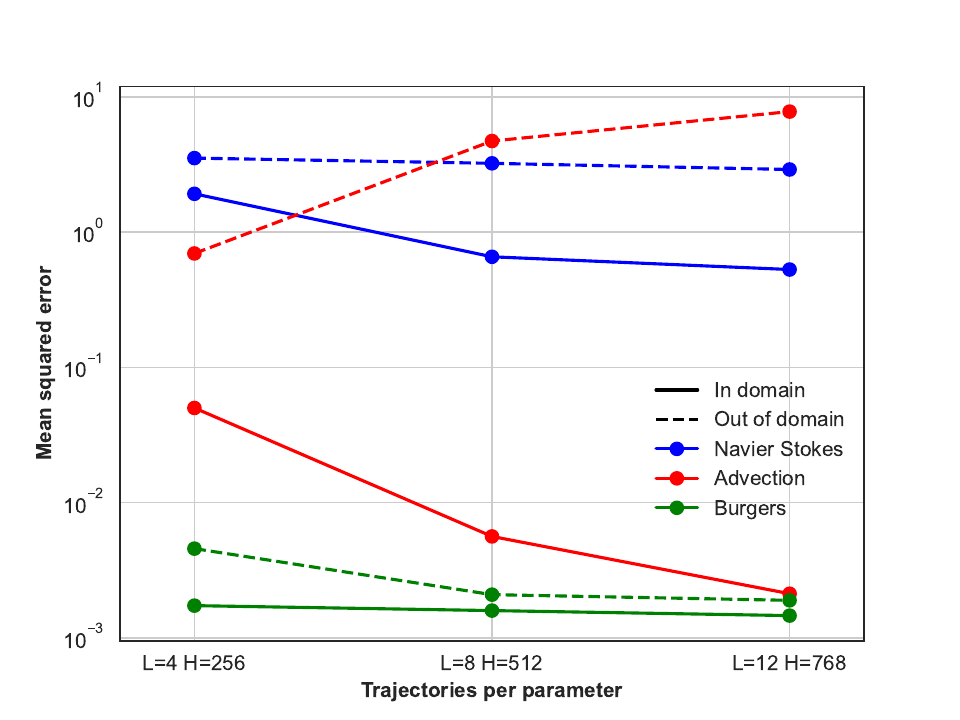}
        \caption{\textbf{In domain and out of domain scaling with model size}: The above plot shows how the in-domain prediction MSE varies with an increase in the model size (i.e. the number of learnable parameters). We observe that the prediction error decreases monotonically with an increase in the model size for all systems in the in domain setting. In the out of domain setting, we observe that the prediction error for Burgers and 2D-NSE decreases steadily with increasing size, however the error for advection actually increases. This anomaly for advection might be due to the larger models overfitting the training data in advection.}
        \label{fig:scaling-model-id}
    \end{figure}


    
    In this experiment we finetune a pretrained PDE\kern+.15em-T, on 100 trajectories per parameter for the Burgers equation (pretraining parameters are same as in Table \ref{table:parameters}). We finetune the model with 100 trajectories for each out of domain parameter for only 10 epochs . We also carry out a similar exercise with a model having randomly initialized weights. In Figure \ref{fig:finetuning} we see that finetuning does help improve the performance of the pretrained model. Performance post finetuning is significantly better than training on the same data from scratch, thus demonstrating the efficacy of pretraining, finetuning methodology.

\subsection*{Scaling with more training data}

    Transformer-based architectures have been shown to improve with more data. We investigate whether the same applies for our approach as well. On progressively increase the training data size (i.e. the number of trajectories per parameter value $\omega$), in Figure \ref{fig:scaling-data-id} we see that the mean squared errors (averaged over all $\omega$) show us interesting trends in the in-domain and out-of-domain settings. The burgers equation scales very well in both settings, and this is not surprising for the chosen system parameters. For high values of dynamic viscosity, the solution for most trajectories diffuses to a constant solution quickly, which the network learns to predict very well.
    For the advection equation we see reasonable scaling in the in-domain setting and some evidence of scaling in the out of domain setting. We believe that the auto-regressive nature of testing, causes error accumulation over long time rollouts which our model has not learned to handle well, due to the multi step input next step prediction approach to training. For the Compressible Navier Stokes we see evidence of scaling in the in-domain setting and that does not seem to be the case for the out-of-domain testing. We believe the reason for this to be similar to what is happening with the advection equation. The model might also possibly be suffering from under fitting for the advection and Navier-Stokes systems and the observed scaling behaviour is combination of both issues.
    
    We carry out a similar analysis for the FNO, details of which can be found in \ref{sec:FNO_scaling}. What we observe is that the FNO learns the relevant information very quickly and does not improve performance with more data, i.e the performance plateaus.

    While we limit our analysis to a maximum of 500 trajectories for each $\omega$ value, from the observed trend and results of the FNO scaling, it is likely that the ViT will continue to improve with more data, and can likely surpass the FNO with sufficient data and compute. 
    
\subsection*{Scaling with larger models}

    Transformer based networks have also been shown to have improved performance with larger model size, i.e. with a larger number of learnable weights. The original ViT \cite{vit} had 12 Transformer layers (L), with 12 self-attention heads (A) per layer and a hidden size (H) of 768. Previous work \cite{well-read-students} explores, among other things, the performance of BERT-like Transformer networks with a variable number of layers and hidden sizes. Working on similar lines, we show the performance of our model with \textit{[L=2, H=128], [L=4, H=256], [L=8, H=512]} in Figure \ref{fig:scaling-model-id}, and we see good scaling in the in-domain setting. In the out of domain-setting while we observe scaling for the Burgers and Advection equation, and we see an anomaly for the Navier-Stokes system which might be due to the larger models overfitting to the training data.

%% file: src/experiments/tables/parameters.tex
\begin{table}[]
    \centering
    \begin{tabular}{@{}c|c|c|c@{}}
    \toprule
     & \textbf{Advection} & \textbf{Burgers} & \textbf{Navier Stokes} \\
    Parameters & [$a$] & [$\nu$] & [($M$, $\zeta$, $\eta$)] \\
    \midrule
    In-domain set & [0.1, 0.4, 2.0] & \begin{tabular}[c]{@{}c@{}} [0.001, 0.004,\\0.01, 0.04,\\0.1, 0.4,\\1.0, 4.0]\end{tabular} & \begin{tabular}[c]{@{}c@{}} [(0.1, 0.01, 0.01),\\ (0.1, 0.1, 0.1),\\ (1.0, 0.1, 0.1)] \end{tabular} \\
    \midrule
    Out-of-domain set & [0.2, 1.0, 4.0] & \begin{tabular}[c]{@{}c@{}} [0.002, 0.02, \\0.2, 2.0] \end{tabular} & [(1.0, 0.01, 0.01)] \\ 
    \bottomrule
    \end{tabular}
    \caption{In-domain (ID) and out of domain (OOD) set of parameters used to train and test the model.}
    \label{table:parameters}
\end{table}

%% file: src/conclusion.tex
\section*{Conclusion}
We show that making use of self supervised pre-training allows us to learn a family of solution operators for a given PDE, thus creating a solver that can reasonably generalize across system parameters.  Similar to other transformers based architectures we see that the PDE\kern+.15em-T is also data hungry and shows good scaling in performance when provided with more data in both in-domain and out-of-domain settings for certain systems. The scaling behaviour for advection and compressible Navier Stokes possibly suffer from error accumulation due to an autoregressive rollout as well as under-fitting to the training data.

We see that the FNO performs exceptionally well on the systems experimented on but are limited in their potential to scale. We also show that finetuning a pre-trained model using small amounts of data allows for improvement of solution predictions, thus suggesting that building large scale pre-trained models might also be the future for the field of neural PDE solvers.

We demonstrate a proof of concept that transformer based architectures with sufficient data and size could approach other state of the art neural PDE solvers.  The current work is restricted to one and two dimensional PDEs, but can be extended to PDEs in three dimensions. In this work we have limited ourselves to a subset of PDEs due to the choice of dataset. In our future work, we will test on 3 dimensional datasets and extend the work to other PDEs.

Scaling behaviour for certain systems can be improved by using models such as \cite{pderefiner} that allow for longer rollouts by using some kind of refinement process.

%% file: src/supplementary.tex
\section{Supplementary information}
\subsection{Data details}
\label{subsec:data_appendix}

As mentioned earlier we use the publicly available PDEBench dataset \cite{PDEBench} to train and test the performance of all the models used in this work. In particular we make use of datasets for the 1D Advection Equation, 1D Burgers Equation and 2D Compressible Navier Stokes systems. Below we list specifics about the dataset of each system:

\subsubsection{1D Advection}

The advection equation models simple linear advective behaviour as described by the following PDE:

\begin{align}
    \frac{\partial u}{\partial t} + a \frac{\partial u}{\partial x} &= 0\\
    u(x, 0) &= u_0(x)
\end{align}

where $\mathrm{a}$ is the advection speed. We use this equation to see how well how model is able to capture simple advective behaviour. The advection speed $a$, is the system parameter of interest. Higher values of $a$, causes information to be propagated faster.

The dataset contains 10,000 trajectories with 1024 grid points from $(0, 1)$ with uniform spacing. Each trajectory has 200 timesteps for $t \in (0, 2]$. Periodic boundary condition was used to create the data.

\subsubsection{1D Burgers}

The Burgers equation is given as:
\begin{align}
    \frac{\partial u}{\partial t} + u\frac{\partial u}{\partial x} &= \nu \frac{\partial^2 u}{\partial x^2} \\
    u(x, 0) &= u_0(x)
\end{align}

where $\nu$ is the dynamic viscosity of the fluid, and the parameter we expect our model to generalize. Training the model on the burgers equation allows us to understand how it captures non linear behaviour (second term on the left hand side) as well as diffusive behaviour (term on the right hand side).

The dataset contains 10,000 trajectories on a uniform grid 1024 nodes in $(0, 1)$. Similar to the linear advection above data is available for 200 timesteps where $t \in (0, 2]$, as well as a periodic boundary condition.

\subsubsection{2D Compressible Navier Stokes}
The compressible Navier Stokes equation is given by:

$$
\begin{aligned}
\partial_t \rho+\nabla \cdot(\rho \mathbf{v}) & =0 \\
\rho\left(\partial_t \mathbf{v}+\mathbf{v} \cdot \nabla \mathbf{v}\right) & =-\nabla p+\eta \triangle \mathbf{v}+(\zeta+\eta / 3) \nabla(\nabla \cdot \mathbf{v}) \\
\partial_t\left[\epsilon+\frac{\rho v^2}{2}\right] & +\nabla \cdot\left[\left(\epsilon+p+\frac{\rho v^2}{2}\right) \mathbf{v}-\mathbf{v} \cdot \sigma^{\prime}\right]=0
\end{aligned}
$$

where $\rho$ is the density, $\mathbf{v}$ is the velocity, $p$ is the pressure, $\epsilon = p/(\Gamma-1)$ is the internal energy, $\Gamma = 5/3$, $\sigma^{\prime}$ is the stress tensor, $\eta$ is the shear viscosity and $\zeta$ is the bulk viscosity. The parameters we hope to generalize across are the Mach number $M = |v|/c$ (where $c=\sqrt{\Gamma p/\rho}$ is the speed of sound), and the shear and bulk viscosities given by $\eta$ and $\zeta$ respectively.

The dataset contains 1000 trajectories on a uniform $128 \times 128$ grid where the domain limits are $(0, 1) \times (0, 1)$. Each trajectory contains 21 uniformly spaced timesteps . The initial velocity field is random with a uniform pressure and density field.

\subsection{Data split}
We used a upto 500 trajectories per parameter value for training (80\%-20\% train-val split), and a fixed set of 100 trajectories for the test sets (for each system).

\subsection{FNO scaling with dataset size}

Similar to the PDE\kern+.15em-T we carry out a study to understand the scaling behaviour of the FNO on increasing the size of the dataset. Compared to the PDE\kern+.15em-T we train the FNO on all the available data from the dataset.

\begin{figure}
        \centering
        \includegraphics[width=\linewidth]{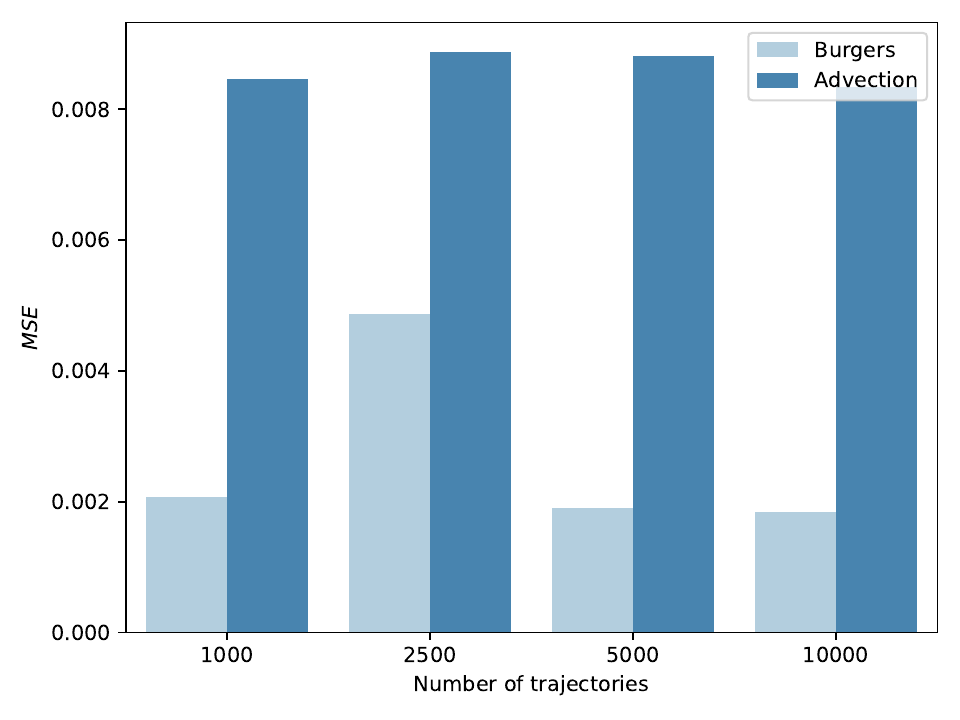}
        \caption{\textbf{FNO Scaling of Advection and Burgers equations:} On scaling the number of trajectories in the dataset for the FNO, we see that the performance remains nearly the same for both the Advection and Burgers equations, suggesting that the FNO learns quickly from little data.}
        \label{fig:FNO_scaling}
    \end{figure}

\begin{figure}
    \centering
    \includegraphics[width=\linewidth]{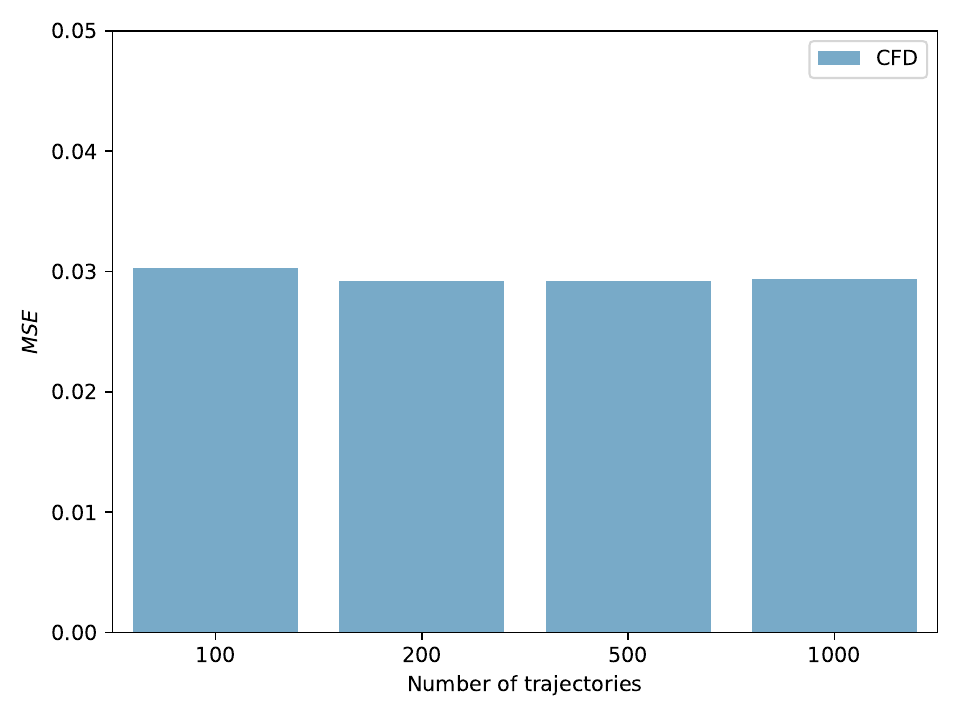}
    \caption{\textbf{FNO Scaling on Compressible Navier Stokes equations}: Scaling the dataset size of the FNO on the Compressible Navier-Stokes dataset, we see a behaviour similar to the Burgers and Advection systems, where the performance remains constant.}
    \label{fig:FNO_CFD}
\end{figure}

Results reported in figures \ref{fig:FNO_scaling} and \ref{fig:FNO_CFD} are averaged across all the parameters that we used to train the PDE\kern+.15em-T.

We see that for the selected hyperparameters of the model learns the relevant information very quickly (with low amounts of data) for all systems and the performance plateaus out.

\label{sec:FNO_scaling}

\subsection{Model details}
\subsection{General details}
We use a patch size of [32, 1] for the Advection and Burgers equations and a patch size of [8, 8, 1] for the 2D Compressible Navier Stokes equation. Unless explicitly specified the transformer contains 4 hidden layers and 4 attention heads per layer, with a hidden dimension of 256. We use Adam as our optimizer.
\subsubsection{Pretraining}
We train the models upto 100 epochs, with a effective batch size of 32. We use reduce the learning rate by a factor of half when the validation loss plateaus, and we start with a learning rate of $10^{-6}$ for Advection, Burgers and $10^{-8}$ for the Navier Stokes case. We make use early stopping to prevent overfitting.

\subsubsection{Finetuning}
We carry out finetuning for 10 epochs with an effective batch size of 32.

\subsubsection{FNO}
The implementation of the FNO used was as provided in \url{https://github.com/pdebench/PDEBench}, setting the training type to "autoregressive".
\begin{itemize}
    \item Epochs: 250
    \item Optimizer: Adam
    \item Learning rate: $1e-3$
    \item Weight decay: $1e-4$
    \item Number of modes: 12
    \item Number of channels for Fourier layer: 20
    \item LR Scheduler: Step scheduler
    \item Scheduler gamma: 0.5
\end{itemize}

\subsection{Software}
To train our model we used PyTorch lightning v1.5.6. The FNO training and testing was carried out using PyTorch v1.13.1.

\subsection{Hardware}
Our model was trained on 4 Nvidia Tesla K80 GPUs using PyTorch lightning's distributed data parallel strategy. The FNO was trained and tested on a single Nvidia A100 GPU.